\documentclass{article}

\usepackage[]{acl}

\usepackage{xcolor} 
\usepackage{booktabs} 
\usepackage{hyperref}       
\usepackage{url}            
\usepackage{amsfonts}       
\usepackage{nicefrac}       
\usepackage{microtype}      
\usepackage{amssymb}
\usepackage{amsbsy}
\usepackage{amsmath}
\usepackage{amsthm}
\usepackage{latexsym}
\usepackage{subcaption}
\usepackage{wrapfig}
\usepackage{graphicx}
\usepackage{tabularx}
\usepackage{ulem}
\usepackage{longtable} 
\usepackage{mathtools}
\usepackage{arydshln}
\usepackage{multirow}
\usepackage{wrapfig,lipsum}
\usepackage{bbm}
\usepackage[most]{tcolorbox}
\usepackage{lmodern}
\usepackage{tikz}
\usepackage{algorithm} 
\usepackage{algorithmic} 
\usepackage{amsmath} 

\usepackage{amsmath,amsfonts,bm}

\def\eqref#1{equation~\ref{#1}}

\def\1{\bm{1}}

\DeclareMathAlphabet{\mathsfit}{\encodingdefault}{\sfdefault}{m}{sl}
\SetMathAlphabet{\mathsfit}{bold}{\encodingdefault}{\sfdefault}{bx}{n}

\DeclareMathOperator*{\argmin}{arg\,min}

\usepackage{hyperref}
\usepackage{url}
\usepackage{pgfplots}
\pgfplotsset{compat=1.18}

\usepackage{times}
\usepackage{latexsym}

\usepackage[T1]{fontenc}
\usepackage[utf8]{inputenc}

\usepackage{microtype}

\usepackage{inconsolata}

\usepackage{graphicx}
\hypersetup{
    colorlinks,
    linkcolor={red!50!black},
    citecolor={blue!50!black},
    urlcolor={blue!80!black}
}

\newcommand{\eg}{\textit{e.g.}}
\newcommand{\ie}{\textit{i.e.}}

\title{Aligning Large Language Models via Fully Self-Synthetic Data}

\author{
  Shangjian Yin\textsuperscript{1} \quad
  Zhepei Wei\textsuperscript{2} \quad
  Xinyu Zhu\textsuperscript{2} \quad
  Wei-Lin Chen\textsuperscript{2} \quad
  Yu Meng\textsuperscript{2} \\
  \textsuperscript{1}University of California, Riverside \quad
  \textsuperscript{2}University of Virginia \\
  \texttt{syin041@ucr.edu} \quad
  \texttt{\{zhepei.wei, xinyuzhu, wlchen, yumeng5\}@virginia.edu}
}

\begin{document}

\maketitle

\begin{abstract}
Traditional reinforcement learning from human feedback (RLHF) for large language models (LLMs) relies on expensive human-annotated datasets, while Reinforcement Learning from AI Feedback (RLAIF) also incurs significant costs, requiring the collection of diverse prompts and corresponding responses, often necessitating external reward models or proprietary models like GPT-4 to annotate preference pairs. 
In this work, we introduce Self-Alignment Optimization (SAO), a fully self-synthetic framework for LLM alignment, where all training data, including prompts (\ie, user queries), responses, and preferences, are generated by the model itself.
Specifically, SAO first instructs the LLM to engage in persona role-play and generate diverse prompts and responses, which are then self-evaluated for preference optimization.
Extensive experiments demonstrate that SAO effectively enhances the model’s chat capabilities on standard benchmarks like AlpacaEval 2.0, while maintaining strong performance on downstream objective tasks (\eg, question-answering, math reasoning).
Our work provides a practical solution for self-improvement in aligning LLMs, and the code for reproducing our results is available at: 
\url{https://github.com/SJY8460/SAO}.
\end{abstract}

\section{Introduction}

Large language models (LLMs) have revolutionized the field of natural language processing (NLP), demonstrating remarkable capabilities in tasks such as mathematical reasoning, code generation, and dialogue generation \citep{cobbe2021training, wei2022chain, bubeck2023sparks, chen2024selfplayfinetuningconvertsweak,yin-etal-2025-eclm,yin-etal-2025-midlm}.

A key advancement in LLMs is their alignment with human preference to create more helpful and reliable assistants~\citep{mishra2021cross, victor2022multitask, chung2022scaling, thoppilan2022lamda}. Common approaches include supervised fine-tuning (SFT) \citep{ouyang2022training, tunstall2023zephyr}, based on human-demonstration pairs, and reinforcement learning from human feedback (RLHF) ~\citep{christiano2017deep, ziegler2019fine, stiennon2020learning, bai2022training}, which leverages signals from human preferences.

However, collecting demonstrations and preference labels is a expensive, time-consuming process, involving substantial human annotating efforts.
To address this challenge, reinforcement learning from AI feedback (RLAIF) has been gaining attention, where a reward model is trained using AI-labeled preference data or directly by LLMs through specially designed prompts to annotate preference pairs~\citep{lee2024rlaifvsrlhfscaling}. 
However, RLAIF remains costly, typically requiring strong, proprietary models (\eg, GPT-4) or specialized reward model designs \citep{jiang2023llmblenderensemblinglargelanguage,ArmoRM} to work effectively.
Moreover, these approaches often incur additional overhead from data filtering to obtain the final clean dataset \citep{xu2024magpiealignmentdatasynthesis}. 

Recent self-improvement approaches for LLM alignment, such as Self-Rewarding~\citep{yuan2024selfrewardinglanguagemodels} and SPPO~\citep{wu2024selfplaypreferenceoptimizationlanguage}, have demonstrated promising results. However, these approaches are not fully self-synthetic and still require external intervention: Self-Rewarding relies on human-labeled data as few-shot templates to generate synthetic data and then mixes both human and synthetic data for model optimization, while SPPO requires prompts from existing datasets and external reward models for preference labeling. Such dependencies limit their scalability and accessibility.

In this paper, we propose a fully self-synthetic method for LLM alignment, termed Self-Alignment Optimization (SAO), eliminating the need for costly training data collection and annotation.
Drawing inspiration from the compress-and-decompress approach to world knowledge from a persona perspective~\citep{ tseng2024talespersonallmssurvey,ge2024incontextautoencodercontextcompression, chan2024scalingsyntheticdatacreation} and the success of self-improvement mechanisms ~\citep{samuel2000some, chen2024selfplayfinetuningconvertsweak}, SAO begins with (1) instructing the LLM to engage in persona role-play and generate diverse prompts (\textit{i.e.}, user queries).
(2) The LLM then generates paired responses for each prompt, which will be ranked through self-judgment.
(3) Lastly, preference optimization is performed to fine-tune the model based on the ranked responses~\citep{amini2024direct, meng2024simposimplepreferenceoptimization}.

Across multiple standard chat benchmarks for LLM alignment, SAO demonstrates substantial performance improvements compared to the backbone model.
For example, on AlpacaEval 2.0, SAO boosts the Length-Controlled Win Rate (LC) and Win Rate (WR) of Gemma-2-9B-it by 18.1\% and 27.9\%, respectively.
On MT-Bench, SAO improves the average score of Gemma-2-9B-it from 8.41 to 8.66 points. On Arena-Hard, SAO also boosts the model's WR from 40.8\% to 54.3\%.

In addition, we find that SAO maintains or even enhances the backbone LLM's performance on downstream objective NLP tasks, though the model was not trained using domain-specific data.
In the evaluation of the Open LLM Leaderboard, SAO enables Gemma-2-9B-it to achieve an average score of 74.41 across all benchmarks, surpassing its baseline score of 74.28, without any further training.
We highlight that this improvement is significant, as models trained on manually crafted datasets often enhance alignment at the cost of compromised general capability~\citep{meng2024simposimplepreferenceoptimization}.
For instance, Gemma-2-9B-it trained on UltraFeedback~\citep{cui2024ultrafeedbackboostinglanguagemodels} achieves only a score of 70.38 using the same training algorithm as SAO. These results highlight SAO’s effectiveness in synthesizing high-quality data that enhances a model’s subject-specific capabilities while preserving its downstream performance in a more data-efficient manner.

To better understand the empirical results, we further conduct an in-depth analysis with the Gemma-2-9B-it model. Specifically, we compare the LLM's self-synthesized prompts with manually crafted prompts from UltraFeedback. Interestingly, we find that synthetic prompts lead to significantly better performance than those from the Ultrafeedback benchmark when followed by the same self-improvement process in SAO (\ie, Step 2 and 3), with a 16.46\% improvement in WR.
This emphasizes the importance of prompt construction for LLM alignment.
We also find that SAO’s effectiveness largely stems from the surprisingly strong self-judging ability of the backbone model, which even surpasses the external reward model ArmoRM-Llama3-8B-v0.1~\citep{ArmoRM} and even GPT4-level feedback when used within the SAO framework. 
This is further confirmed by the findings in Section~\ref{sec:criteria_design}, which show that the model is robust to various judgment criteria designs, indicating that the model's strong evaluation ability,
thereby contributing to the effectiveness of SAO.

Our contributions are summarized as follows: (1) We propose a self-alignment optimization framework, which aligns the model with its own generated prompts and feedback as preferences, eliminating the need for expensive data collection and annotation.
(2) On multiple standard chat benchmarks, the LLMs trained with SAO achieve competitive results compared to strong proprietary models like GPT-4 and state-of-the-art open-weight models trained on expensively annotated external datasets. (3) In contrast to previous resource-intensive approaches, SAO enhances LLM alignment while maintaining strong downstream objective task performance, providing a practical and scalable solution for improving a model’s chat abilities in a more balanced and effective manner.
(4) We demonstrate that the SAO framework exhibits long-term scaling properties in both synthetic dataset size and iterative optimization, providing a sustainable path for continued model improvement.

\section{Related Work}

\subsection{Synthetic Data for LLMs}
In the context of fine-tuning LLMs, human-crafted data has proven remarkably effective, significantly enhancing performance on tasks like code generation~\citep{roziere2023code, yang2023decoding} and mathematical reasoning~\citep{yuan2023scaling, luo2023wizardmath, zhu2025surprising}. While human-generated data is typically of high quality, acquiring sufficient amounts is often prohibitively expensive. Consequently, the use of synthetic data has gained popularity as a cost-effective substitute for human data. This approach primarily leverages advanced LLMs, such as the GPT series~\citep{radford2019language, brown2020language, openai2023gpt4}, to generate high-quality data~\citep{josifoski2023exploiting, alpaca, vicuna2023, li2023textbooks}.
Recent studies have also emphasized the benefits of using LLMs' rephrasing capabilities to improve prompt responses~\citep{deng2023rephrase, prasad2023rephrase}, as well as augmenting synthetic data for more effective fine-tuning~\citep{yu2023metamath, liu2023tinygsm}. Unlike prior research, which typically relies on more advanced models for generating synthetic training data during, our approach directly generates synthetic data from the target model itself, streamlining the process and reducing dependency on external resources.

\subsection{Persona Roleplay}
Persona roleplay in LLMs can be viewed as a compress-and-decompress mechanism for world knowledge. In this framework, world knowledge is compressed into distributed representations, which are then decompressed by various personas to generate texts based on their unique knowledge~\citep{delétang2024languagemodelingcompression, ge2024incontextautoencodercontextcompression}. Recently, \citet{chan2024scalingsyntheticdatacreation} introduced the \textit{Persona Hub}, a system automatically constructed from vast web data. This hub enables the exploration of most perspectives embedded within LLMs, facilitating the creation of diverse synthetic data at scale, without the need for seed corpora.
In this work, we directly leverage personas from this hub, which can be seamlessly integrated into customized data synthesis prompts, leveraging the LLM's strong role-playing capabilities and offering exceptional versatility.

\subsection{LLM-as-a-Judge}
Using LLM-as-a-Judge prompting to evaluate language models has become a standard approach~\citep{dubois2024alpacafarmsimulationframeworkmethods, AlpacaEval, fernandes2023devil, bai2023benchmarking, saha2023branch, chen2025llm, wei2025instructrag}. This technique is not only employed for evaluation but also for training reward models and curating data, as mentioned in prior works~\citep{lee2023rlaif, chen2023alpagasus, li2023self, wei2025truthrl}. While some studies focus on creating training data to enhance an LLM's performance as a judge \citep{kim2023prometheus, yuan2024selfrewardinglanguagemodels}, our approach uniquely integrates this judging capability with general instruction-following skills, setting it apart from existing methods.

\begin{algorithm}[t]
\caption{Self-Alignment Optimization (SAO)}
\label{alg:self_alignment}
\begin{algorithmic}
\REQUIRE Base model $\mathcal{M}_{\theta_0}$, number of personas $n$, preference optimization algorithm $\mathcal{A}$
\ENSURE Optimized model $\mathcal{M}_{\theta_1}$
\STATE Initialize personas templates $\{r_1, r_2, \ldots, r_n\}$
\STATE Initialize dataset $\mathcal{D} \leftarrow \emptyset$
\FOR{$i = 1, 2, \ldots, n$}
    \STATE $x_{\text{prompt}}^i \leftarrow \mathcal{M}_{\theta_0}(r_i)$ \hfill  \S~\ref{subsec:prompt_generation}
    \STATE $y_1, y_2 \leftarrow \mathcal{M}_{\theta_0}(\cdot \mid x_{\text{prompt}}^i)$ \hfill  \S~\ref{subsec:response_generation}
    \STATE $(y_{\text{w}}, y_{\text{l}}) \leftarrow \mathcal{R}_{\theta_0}(y_1, y_2 \mid x_{\text{prompt}}^i, x_{\text{rank}})$ \hfill  \S~\ref{subsec:self_judge}
    \STATE $\mathcal{D} \leftarrow \mathcal{D} \cup \{(x_{\text{prompt}}^i, y_{\text{w}}, y_{\text{l}})\}$ \hfill  \S~\ref{subsec:construct}
\ENDFOR
\STATE Optimize: $\theta_1 \leftarrow \argmin_{\theta} \mathcal{L}_{\mathcal{A}}(\mathcal{M}_{\theta})$
\STATE \textbf{Return} $\mathcal{M}_{\theta_1}$
\end{algorithmic}
\end{algorithm}

\section{Self-Alignment Optimization}
We present the overview of Self-Alignment Optimization (SAO) in Algorithm \ref{alg:self_alignment} and elaborate on its design in the following sections.

\subsection{Diverse Prompt Generation}\label{subsec:prompt_generation}

To facilitate a comprehensive range of training scenarios, we utilize an LLM denoted as $\mathcal{M}$, parameterized by $\theta$, for the generation of diverse prompts tailored to specific persona roles, as depicted in Figure \ref{fig:eval_prompt} (top). Given a set of role templates $\mathcal{R} = \{r_i\}_{i=1}^n$, we derive a unique prompt for each persona:
\begin{equation}
    x_{\text{prompt}}^i = \mathcal{M}_\theta(r_i)
\end{equation}

In this context, $x_{\text{prompt}}^i$ represents the prompt generated for the $i$-th persona $r_i$. To ensure the diversity of generated prompts, we impose a constraint such that each persona can generate only a single question. The persona resources are randomly sampled from Persona-Hub \citep{chan2024scalingsyntheticdatacreation}.

\subsection{Pairwise Response Generation}\label{subsec:response_generation}

For each generated prompt, we create a pair of responses to enable comparative evaluation. Let $\mathcal{X}$ be the space of prompts and $\mathcal{Y}$ the space of responses. For each prompt $x_{\text{prompt}}^i \in \mathcal{X}$, we generate two responses $y_1, y_2 \in \mathcal{Y}$ using the $\mathcal{M}_\theta$:
\begin{equation}
    y_1, y_2 \sim \mathcal{M}_\theta(\cdot \mid x_{\text{prompt}}^i)
\end{equation}

Generating additional responses could potentially yield better performance but would increase computational costs and evaluation time. We leave this exploration for future work.

\subsection{Self-Judgment}\label{subsec:self_judge}

To assess the quality of generated responses, we implement a self-judgment mechanism. This process entails the LLM evaluating its own outputs, thereby simulating human preferences. As illustrated in Figure \ref{fig:eval_prompt} (bottom), we query the LLM with a ranking prompt $x_{\text{rank}}$ to compare the responses $y_1$ and $y_2$ based on their relevance and quality relative to $x_{\text{prompt}}^i$:
\begin{equation}
    (y_{\text{w}}, y_{\text{l}}) = \mathcal{R}_\theta(y_1, y_2 \mid x_{\text{prompt}}^i, x_{\text{rank}})
\end{equation}

Here, $y_{\text{w}}$ and $y_{\text{l}}$ represent the superior and inferior responses, respectively. The function $\mathcal{R}_\theta$ encapsulates the LLM's decision-making process in ranking the responses.

\subsection{Dataset Construction}\label{subsec:construct}

We construct a synthetic dataset $\mathcal{D}$ by aggregating the generated prompts and ranked responses for each persona:
\begin{equation}
    \mathcal{D} = \{(x_{\text{prompt}}^i, y_{\text{win}}^i, y_{\text{lose}}^i)\}_{i=1}^n
\end{equation}

where $n$ is the total number of personas. This dataset forms the cornerstone of our preference optimization process, allowing the model to learn from its own generated and ranked responses across diverse personas.

\subsection{Preference Optimization}\label{subsec:optimization}

Recent advancements in preference optimization have demonstrated significant potential in aligning LLMs with human preferences. Techniques such as Direct Preference Optimization (DPO) \citep{Rafailov2023DirectPO} and Simple Preference Optimization (SimPO) \citep{meng2024simposimplepreferenceoptimization} have gained prominence due to their efficacy in fine-tuning LLMs to better reflect human preferences.
In this study, we employ SimPO due to its suitability for our dataset, which frequently contains longer responses. Its length normalization technique effectively captures nuanced information at the token level, making it particularly well-suited to our requirements and we provide a more detailed analysis and comparison of these methods in Section \ref{different_algorithms}.

SimPO introduces a length-normalized reward formulation that aligns with the likelihood metric guiding generation with a scaling constant $\beta$:
\begin{equation}
    \begin{split}
    r(x,y) &= \frac{\beta}{|y|} \log \mathcal{M}_\theta(y \mid x) \\
    &= \frac{\beta}{|y|} \sum_{i=1}^{|y|} \log \mathcal{M}_\theta(y_i \mid x,y_{<i})
    \end{split}
\end{equation}

Additionally, it incorporates a target reward margin $\gamma > 0$ to ensure a minimum difference between the rewards of winning and losing responses:
\begin{equation}
    p(y_w \succ y_l \mid x) = \sigma \left( r(x, y_w) - r(x, y_l) - \gamma \right)
\end{equation}

The overall objective is then formulated as:
\begin{equation}
    \label{eq:simpo}
    \begin{split}
    \mathcal{L}(\mathcal{M}_\theta) = &- \mathbb{E}_{(x, y_w, y_l) \sim \mathcal{D}}\biggl[ 
    \log \sigma \biggl( \frac{\beta}{|y_w|}  \log  \mathcal{M}_\theta(y_w|x) \\
    &\hspace{2em} - \frac{\beta}{|y_l|} \log \mathcal{M}_\theta(y_l|x) - \gamma \biggr) \biggr]
    \end{split}
\end{equation}

It guides the optimization process, enabling the model to learn from self-generated preferences and improve the alignment with desired outcomes.

\begin{table*}[t]
\label{tab:result-alpaca}
\centering
\caption{Comparative analysis of various baseline models and our proposed SAO method using AlpacaEval 2.0. The table presents Length-Controlled Win Rate (LC), Win Rate (WR), and Standard Deviation (STD) for each model, evaluated against GPT-4-Turbo-1106 and Qwen2-72B-Instruct. We also provide additional judge analysis in Section~\ref{Judgers}.}
\resizebox{0.9\textwidth}{!}{%
\begin{tabular}{l c c c c c c}
\toprule
\multirow{3}{*}{Model} & \multicolumn{6}{c}{AlpacaEval 2.0} \\
\cmidrule{2-7}
& \multicolumn{3}{c}{GPT-4-Turbo-1106} & \multicolumn{3}{c}{Qwen2-72B-Instruct} \\
\cmidrule(lr){2-4} \cmidrule(lr){5-7}
& LC (\%) & WR (\%) & STD & LC (\%) & WR (\%) & STD \\
\midrule
\multicolumn{7}{l}{\textit{Off-the-shelf Models}} \\
\midrule
Llama-3-8B-Instruct & 22.9 & 22.6 & 1.3 & 29.4 & 29.2 & 1.6 \\
Yi-34B-Chat & 27.2 & 29.7 & 1.3 & 33.3 & 37.0 & 1.7 \\
GPT-4-Turbo-04-09  & 55.0 & 46.1 & 1.5 & 49.0 & 39.1 & 1.7 \\
Gemma-2-9B-it & 51.1 & 38.1 & - & 56.5 & 39.3 & 1.7 \\
Claude-3.5-Sonnet & 52.4 & 40.6 & 1.5 & \textbf{56.8} & 40.5 & 1.7 \\
GPT-4o-05-13 & \textbf{57.5} & \textbf{51.3} & 1.5 & 51.8 & \textbf{44.7} & 1.8 \\
\midrule
\multicolumn{7}{l}{\textit{Models Trained Using External Generated/Labeled Dataset}} \\
\midrule
Self-Rewarding-70B-Iter3 \citep{yuan2024selfrewardinglanguagemodels} & -	& 20.4 & - & - & - & - \\
Llama-3-Instruct-8B-SimPO \citep{meng2024simposimplepreferenceoptimization} & 53.7 & 47.5 & - & 54.2 & 45.9 & 1.8 \\
Gemma-2-9B-SPPO-Iter3 \citep{wu2024selfplaypreferenceoptimizationlanguage} & 53.3	& 47.8 & - & - & - & - \\
Llama-3-8B-Magpie-SFT-v0.1 \citep{xu2024magpiealignmentdatasynthesis} & 24.2 & 25.2 & - & 26.2 & 29.2 & 1.6 \\
Gemma-2-9B-it-SimPO \citep{meng2024simposimplepreferenceoptimization} & \textbf{72.4} & \textbf{65.9} & 1.4 & \textbf{74.5} & \textbf{65.5} & 1.7 \\

\midrule
\multicolumn{7}{l}{\textit{Models Trained Only Using Self-Synthetic Dataset}} \\
\midrule

Llama-3-Instruct-8B-SAO (Ours) & 33.3 (+10.4) & 39.0 (+16.4) & 1.4 & 42.3 (+12.9) & 49.1 (+19.9) & 1.8 \\
Gemma-2-9B-it-SAO (Ours) & \textbf{69.2} (+18.1) & \textbf{66.0} (+27.9) & 1.4 & \textbf{76.0} (+19.5) & \textbf{71.6} (+32.3) & 1.6 \\
\bottomrule
\end{tabular}%
}
\end{table*}

\section{Experiments}

\subsection{Experiment Setting}

In our experiments, we use the Gemma-9B-it model as the base and apply a similar fine-tuning process to Llama-3-8B-Instruct. To ensure diverse prompts and responses, we set the temperature to 0.6 and utilize VLLM for accelerated generation. For preference optimization, we incorporate Flash Attention 2 and bfloat16 precision, with hyperparameters set to $\beta = 10$ and $\gamma = 3$. Additionally, we employ DeepSpeed with ZeRO-3 optimization for effective memory management and scalability. All experiments are conducted over a single epoch with a global batch size of 128 across four A100 GPUs. The learning rate is set to $1 \times 10^{-6}$, following a cosine decay scheduler with a warmup ratio of 0.1. We use a synthetic dataset of 60k samples as the default setting for self-alignment optimization.

\subsection{Evaluation Metrics and Baselines}

\noindent{\bf Evaluation Metrics.} Our experimental evaluation employs a comprehensive set of metrics to assess model performance across various dimensions. For subjective benchmarks, we primarily focus on AlpacaEval 2 \citep{AlpacaEval}, an LLM-based automatic evaluation benchmark utilizing prompts from AlpacaFarm \citep{dubois2024alpacafarmsimulationframeworkmethods}. In this benchmark, model responses and GPT-4-Turbo generated reference responses are evaluated by GPT-4-Turbo or Qwen2-72B-Instruct annotators. We also incorporate GPT-4 to evaluate two additional subjective benchmarks: Arena-Hard~\citep{arenahard2024}, an automatic evaluation tool featuring 500 challenging user queries, and MT-Bench~\citep{zheng2023judging}, a set of 80 high-quality multi-turn open-ended questions covering topics such as writing, role-playing, math, and coding.
For objective benchmarks, we utilize the Open LLM Leaderboard \citep{open-llm-leaderboard}, which comprises six datasets focusing on various aspects of language model evaluation, including math problem-solving, language understanding, human falsehood mimicking, and reasoning. We adhere to the standard evaluation process, using in-context learning to prompt the models and compute the average score across these six datasets to measure performance comprehensively.

\noindent{\bf Baselines.} In our comparisons, we include a diverse set of baselines. These encompass vanilla models such as GPT-4o-05-13, Claude-3.5-Sonnet, and GPT4-Turbo-04-09. Additionally, we evaluate models trained on external labeled datasets, like Llama-3-Instruct-8B-SimPO \citep{meng2024simposimplepreferenceoptimization}, which has been fine-tuned using the Ultrafeedback dataset \citep{cui2024ultrafeedbackboostinglanguagemodels} for preference optimization.We also consider Self-Rewarding-70B-Iter3 \citep{yuan2024selfrewardinglanguagemodels}, which is trained using a mixture of external labeled datasets and synthetic data. 
Additionally, we compare against the recently developed data-distillation baseline, Llama-3-8B-Magpie-SFT-v0.1 \citep{xu2024magpiealignmentdatasynthesis}, which was originally trained on synthetic SFT pair data generated by the instruction-tuned model itself to improve its base model alignment capability.
Furthermore, we examine Gemma-2-9B-SPPO-Iter3 \citep{wu2024selfplaypreferenceoptimizationlanguage}, which generates responses based on Ultrafeedback prompts and utilizes preference pairs labeled by an external reward model.

\subsection{Main Results}
\label{subsec:results}

\paragraph{Performance on AlpacaEval 2.0.}
Our SAO-tuned models demonstrate substantial performance improvements on AlpacaEval 2.0 when evaluated by both GPT-4-Turbo-1106 and Qwen2-72B-Instruct. When assessed by GPT-4-Turbo-1106, Gemma-2-9B-it-SAO achieves a LC of 69.2\% and a WR of 66.0\%, representing increases of 18.1\% and 27.9\% respectively over the baseline Gemma-2-9B-it (51.1\% LC, 38.1\% WR). This performance surpasses all vanilla models, including the top-performing GPT-4o (05-13) at 57.5\% LC and 51.3\% WR. Moreover, Gemma-2-9B-it-SAO competes closely with models trained on external datasets, approaching the performance of Gemma-2-9B-it-SimPO (72.4\% LC, 65.9\% WR). Similarly, Llama-3-Instruct-8B-SAO exhibits significant improvements, reaching 33.3\% LC and 39.0\% WR, increases of 10.4\% and 16.4\% respectively over the baseline Llama-3-8B-Instruct (22.9\% LC, 22.6\% WR). When evaluated by Qwen2-72B-Instruct, Gemma-2-9B-it-SAO continues to excel, achieving 76.0\% LC and 71.6\% WR. These represent substantial improvements of 19.5\% and 32.3\% over the baseline Gemma-2-9B-it (56.5\% LC, 39.3\% WR) and even outperform models trained on external datasets, such as Gemma-2-9B-it-SimPO (74.5\% LC, 65.5\% WR). Llama-3-Instruct-8B-SAO also demonstrates significant improvement when evaluated by Qwen2-72B-Instruct, reaching 42.3\% LC and 49.1\% WR, increases of 12.9\% and 19.9\% over Llama-3-8B-Instruct (29.4\% LC, 29.2\% WR).

These results underscore the efficacy of our SAO method in enhancing model performance across different base models and evaluation metrics. Notably, SAO achieves these improvements without relying on external labeled datasets, highlighting its potential for efficient and scalable model enhancement.

\paragraph{Performance on MT-Bench and Arena-Hard.}
Our evaluation extended to two other mainstream subjective benchmarks, MT-Bench and Arena-Hard, yielding compelling results that underscore the efficacy of SAO fine-tuning. As shown in Table~\ref{tab:mt-all}, on the MT-Bench benchmark, Gemma-2-9B-it-SAO achieved an average score of 8.66, surpassing the backbone model Gemma-2-9B-it by 0.25 points. In contrast, LLaMA-3-8B-Instruct-SAO maintained its average score of 7.84. 
The Arena-Hard benchmark revealed even more substantial performance gains, with Gemma-2-9B-it experiencing a remarkable increase in WR from 40.8\% to 54.3\% after SAO tuning, marking a 13.5 percentage point improvement. Similarly, LLaMA-3-Instruct-8B's WR rose from 20.6\% to 28.1\%, reflecting a 7.5 percentage point increase. 

These significant advancements across the MT-Bench and Arena-Hard benchmarks highlight the potential of SAO tuning in enhancing model performance on multi-turn, open-ended questions and other diverse and challenging tasks.

\paragraph{Downstream task performance.}
Since the SAO framework typically generates instruction data pairs without ground-truth-style data (e.g., math and code), it is essential to assess its influence on downstream objective task performance. As detailed in Table \ref{tab:result-benchmarks}, we conducted a comprehensive evaluation across diverse tasks using the Open LLM Leaderboard benchmarks. The results demonstrate that SAO-tuned models generally maintain or slightly improve their capabilities compared to their baseline counterparts.

For the Gemma-2-9B series, the SAO-tuned version achieves an average score of 74.41 across all benchmarks, marginally surpassing the baseline Gemma-2-9B-it (74.28). Notably, Gemma-2-9B-it-SAO shows improvements in ARC (+0.42), TruthfulQA (+2.61), and HellaSwag (+0.80), while maintaining comparable performance in other tasks. Similarly, Llama-3-8B-Instruct-SAO (68.20) slightly outperforms its baseline (68.19), with notable enhancements in ARC (+1.62) and HellaSwag (+0.18).

Interestingly, models optimized with externally labeled datasets like Ultrafeedback, such as Gemma-2-9B-it-SimPO and Llama-3-8B-Instruct-SimPO, while achieving impressive results on alignment tasks, show a decrease in overall performance across these general benchmarks. Gemma-2-9B-it-SimPO's average score (70.38) is significantly lower than both the baseline and SAO-tuned versions, with notable declines in Winograd (-4.34) and HellaSwag (-15.08). Llama-3-8B-Instruct-SimPO, despite improvements in specific areas like TruthfulQA (+12.16), also shows a slight overall decrease (67.73) compared to its baseline, primarily due to a substantial drop in GSM8K performance (-20.01).

We hypothesize that this performance discrepancy stems from the nature of externally annotated datasets, which may not align perfectly with the current capabilities of these language models. While such datasets can yield improvements in specific alignment tasks, they may inadvertently compromise the model's general abilities. In contrast, the SAO method, which utilizes self-generated subjective instruction data, appears to more accurately represent and enhance the model's intrinsic capabilities, leading to consistent performance across a wide range of tasks without significant trade-offs.

\begin{table}[t]
    \centering
\caption{Performance on MT-Bench and Arena-Hard benchmarks. MT-Bench shows the lowest agreement with Chatbot Arena compared to AlpacaEval 2.0 and Arena-Hard \citep{meng2024simposimplepreferenceoptimization}, but is included due to its wide adoption in prior work.}
    \label{tab:mt-all}
    \resizebox{0.88\columnwidth}{!}{
        \begin{tabular}{lcc}
            \toprule
            \textbf{Model} & \textbf{MT-Bench} & \textbf{Arena-Hard} \\
            \midrule
            Llama-3-8B-instruct       & 7.84 & 20.6 \\
            \quad + SAO   & 7.84\,(+0.00) & 28.1\,(+7.5) \\
            \midrule
            Gemma-2-9B-it             & 8.41 & 40.8 \\
            \quad + SAO         & 8.66\,(+0.25) & 54.3\,(+13.5) \\
            \bottomrule
        \end{tabular}
    }
\end{table}

\begin{table*}[t]
    \centering
    \caption{Performance comparison of models on downstream NLP benchmarks from the Open LLM Leaderboard. The values in parentheses indicate the number of few-shot examples (shots).}
    
    \resizebox{1.0\textwidth}{!}{%
    \begin{tabular}{lccccccc}
    \toprule
\textbf{Model} & \textbf{ARC (25)} & \textbf{TruthfulQA (0)} & \textbf{Winograd (5)} & \textbf{GSM8K (5)} & \textbf{HellaSwag (10)} & \textbf{MMLU (5)} & \textbf{Average} \\
    \midrule
    Gemma-2-9B-it-SAO & 71.50 & 62.76 & 77.35 & 80.29 & 82.53 & 72.02 & \textbf{74.41} \\
    Gemma-2-9B-it-SimPO & 69.11 & 59.00 & 73.72 & 81.96 & 66.65 & 71.82 & 70.38 \\
    Gemma-2-9B-it & 71.08 & 60.15 & 78.06 & 82.34 & 81.73 & 72.30 & 74.28 \\
    \midrule
    Llama-3-8B-Instruct-SAO & 63.57 & 49.58 & 74.66 & 76.72 & 78.96 & 65.72 & \textbf{68.20} \\
    Llama-3-8B-Instruct-SimPO & 66.64 & 63.86 & 74.74 & 55.65 & 78.97 & 66.51 & 67.73 \\
    Llama-3-8B-Instruct & 61.95 & 51.70 & 75.30 & 75.66 & 78.78 & 65.72 & 68.19 \\
    \bottomrule
    \end{tabular}%
    }
    \label{tab:result-benchmarks}
\end{table*}

\setlength{\tabcolsep}{4pt} 

\begin{figure}[t]
    \centering
    \scalebox{0.9}{  
        \includegraphics[width=\linewidth]{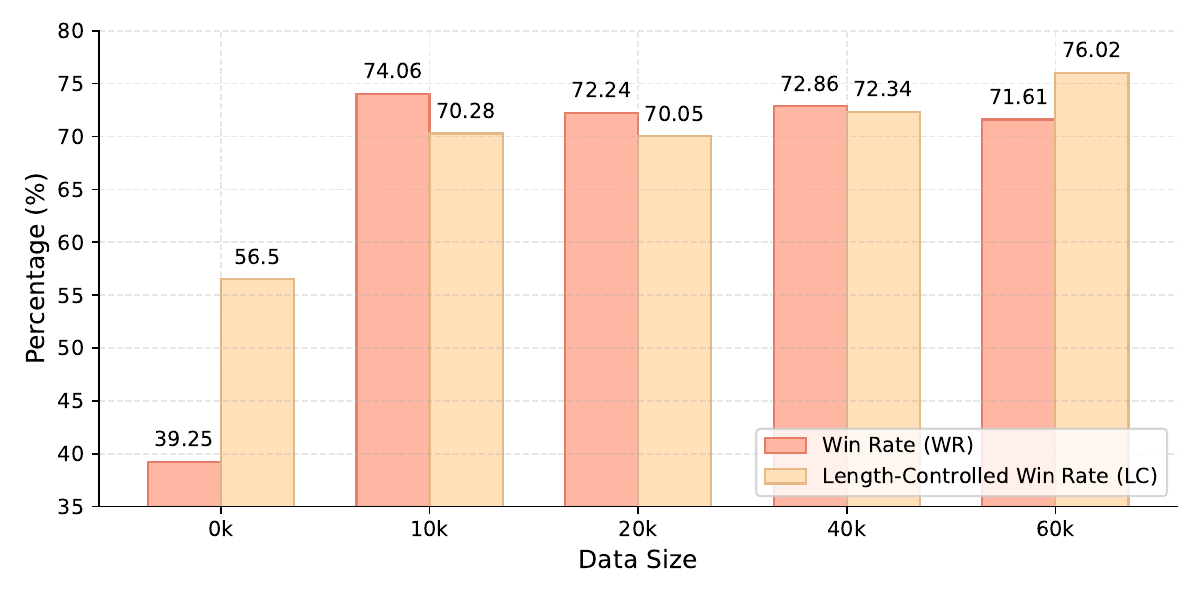}
    }
    \caption{Impact of dataset size on model performance.}
    \label{fig:dataset_size_impact}
\end{figure}

\subsection{Analysis and Ablations}
In this subsection, we primarily utilize Gemma-2-9B-it-SAO to investigate various factors influencing model performance on AlpacaEval 2.0, leveraging Qwen2-72B-Instruct as the evaluation judge. Our observations indicate that Qwen2-72B-Instruct achieves evaluation results comparable to GPT-4, making it a cost-effective alternative. This evaluation includes analyzing the scaling law of the synthetic dataset size. Additionally, for other factors such as optimization algorithms, persona prompt generation, and judging methods, we observed that even with a 10k synthetic dataset, the SAO-tuned model demonstrated promising improvements. Therefore, we focused on these aspects using the 10k synthetic dataset to enhance the efficiency and cost-effectiveness of the evaluation process.

\paragraph{Scaling law of synthetic dataset size.}
As illustrated in Figure \ref{fig:dataset_size_impact}, the performance of Gemma-2-9B-it-SAO improved significantly with an increase in synthetic dataset size. The WR rose from 39.25\% for the vanilla model to 74.06\% with a 10k dataset, stabilizing around 72\% for larger datasets. Similarly, the LC metric reached 76.02\% with a 60k dataset. Interestingly, we observed that even a small amount of self-alignment data can significantly enhance model alignment performance. However, the model's performance appears to saturate after 10k examples, suggesting diminishing returns with further increases in dataset size. 

We hypothesize that more complex prompts generated in the post-SAO stage could further unlock the potential of the model, enabling additional performance gains. Future work could explore innovative prompt generation techniques or incorporate domain-specific nuances to overcome this saturation and maximize the model’s capabilities.

\begin{figure}[t]
\centering
\begin{tikzpicture}
\begin{axis}[
    width=0.88\columnwidth,
    height=5.0cm,
    xlabel={Iteration},
    ylabel={Performance (\%)},
    xmin=-0.6, xmax=2.7,
    ymin=25, ymax=95,
    xtick={0,1,2},
    ytick={30,40,50,60,70,80,90},
    legend pos=south east,
    legend style={font=\small, draw=black!50, fill=white},
    ymajorgrids=true,
    grid style={dashed, gray!20},
    mark size=3.5pt,
]

\addplot[
    color=blue!80!black,
    mark=square*,
    line width=2pt,
    ]
    coordinates {
    (0,56.50)(1,70.28)(2,74.13)
    };
\addlegendentry{LC}

\addplot[
    color=red!80!black,
    mark=triangle*,
    line width=2pt,
    ]
    coordinates {
    (0,39.30)(1,74.06)(2,86.46)
    };
\addlegendentry{WR}

\node[font=\footnotesize, blue!80!black, fill=white, inner sep=1.5pt, anchor=east] 
    at (axis cs:-0.2,56.5) {56.5};
\node[font=\footnotesize, blue!80!black, fill=white, inner sep=1.5pt, anchor=north east] 
    at (axis cs:1.05,68.5) {70.28};
\node[font=\footnotesize, blue!80!black, fill=white, inner sep=1.5pt, anchor=west] 
    at (axis cs:2.15,74.13) {74.13};

\node[font=\footnotesize, red!80!black, fill=white, inner sep=1.5pt, anchor=east] 
    at (axis cs:-0.2,39.3) {39.3};
\node[font=\footnotesize, red!80!black, fill=white, inner sep=1.5pt, anchor=south west] 
    at (axis cs:0.92,75.5) {74.06};
\node[font=\footnotesize, red!80!black, fill=white, inner sep=1.5pt, anchor=west] 
    at (axis cs:2.15,86.46) {86.46};

\end{axis}
\end{tikzpicture}
\caption{Impact of iterative optimization.}
\label{fig:sao_iteration}
\end{figure}
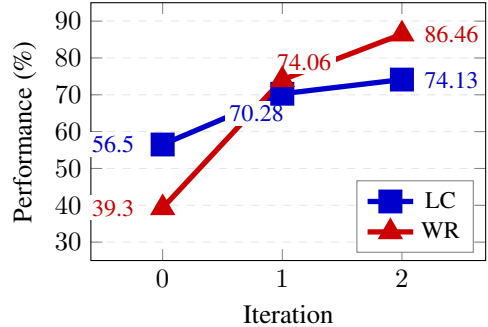

\paragraph{Iterative self-optimization.}

We also explore iterative optimization, where the improved model generates new preference data to further train itself. This shows promising scalability, and we consider future work on balancing dataset size and iteration count for optimal cost-effectiveness.
As shown in Figure~\ref{fig:sao_iteration}, both LC and WR metrics demonstrate consistent improvement across iterations. Notably, the WR metric exhibits substantial gains, jumping from 39.30\% in the vanilla model to 74.06\% after the first iteration, and further increasing to 86.46\% in the second iteration. The LC metric also shows steady growth from 56.50\% to 74.13\%. These results demonstrate the effectiveness of iterative self-optimization in enhancing model alignment capabilities.

\begin{table}[!tbh]
    \centering
    \caption{Experiments with different preference optimization algorithms using same generated dataset.}
    \label{fig:alpaca-eval}
    \scalebox{0.95}{
    \begin{tabular}{lc}
        \toprule
        \textbf{Algorithms} & \textbf{Win Rate (\%)} \\
        \midrule
        None   & 39.25 \\
        DPO                    & 49.81\\
        ORPO                   & 67.33 \\
        \midrule
        SimPO (default)        & 74.04 \\
        \bottomrule
    \end{tabular}
    }
\end{table}

\paragraph{Different optimization algorithms.}
\label{different_algorithms}
To investigate the influence of different optimization algorithms, we compared three mainstream approaches: DPO \citep{xu2024dpo}, ORPO \citep{hong2024orpomonolithicpreferenceoptimization}, and SimPO \citep{meng2024simposimplepreferenceoptimization}. Table \ref{fig:alpaca-eval} illustrates the performance of these algorithms. Starting from the baseline Gemma-2-9B-it model (39.25\% WR), we observed progressive improvements: DPO raised the WR to 49.81\%, ORPO increased it further to 67.33\%, and SimPO achieved the highest WR of 74.04\%. 

\begin{figure}[t]
    \centering
    \scalebox{0.95}{
    \begin{tabular}{c}
        \includegraphics[width=0.49\textwidth]{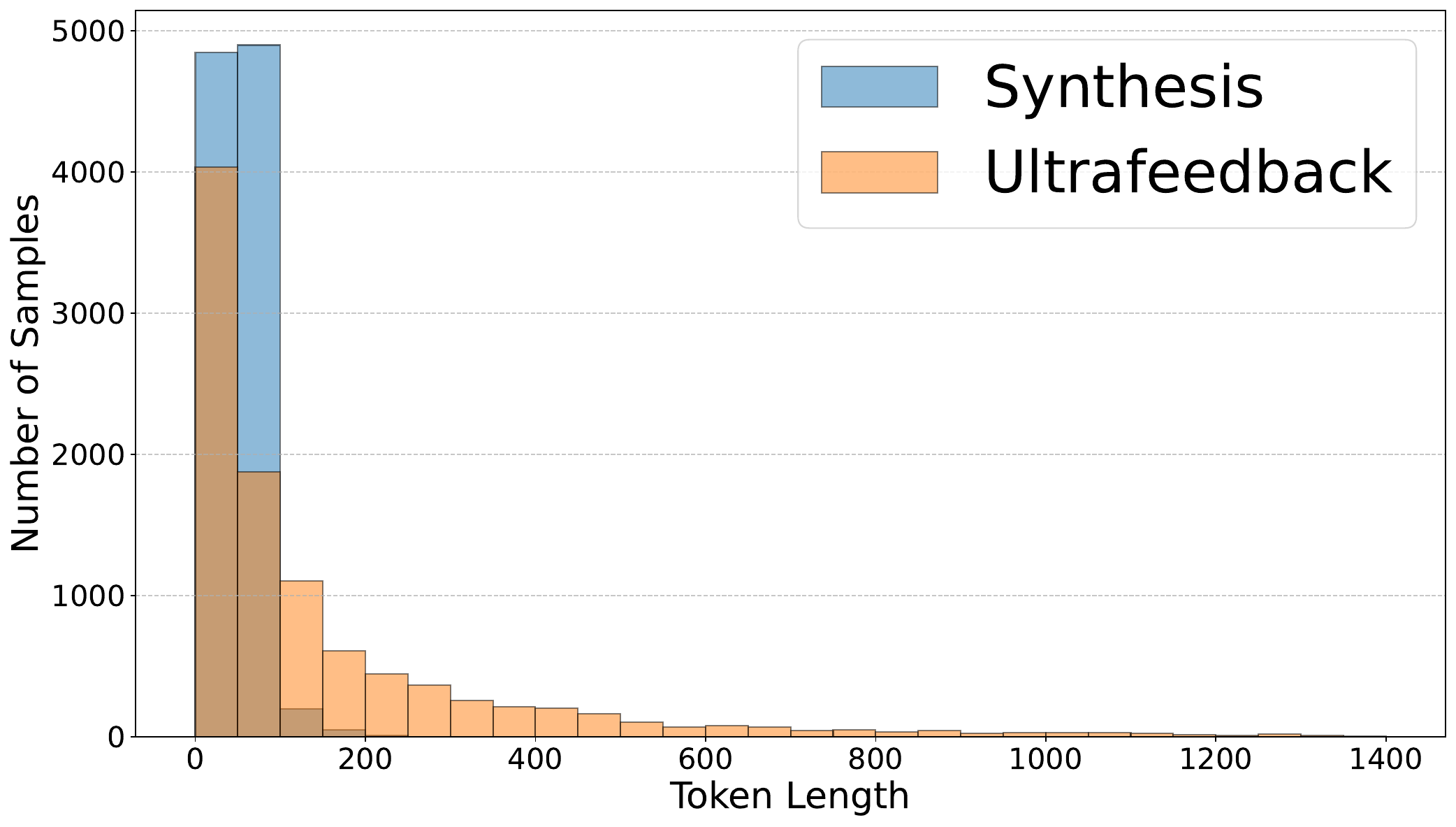} \\[1em]
        \includegraphics[width=0.49\textwidth]{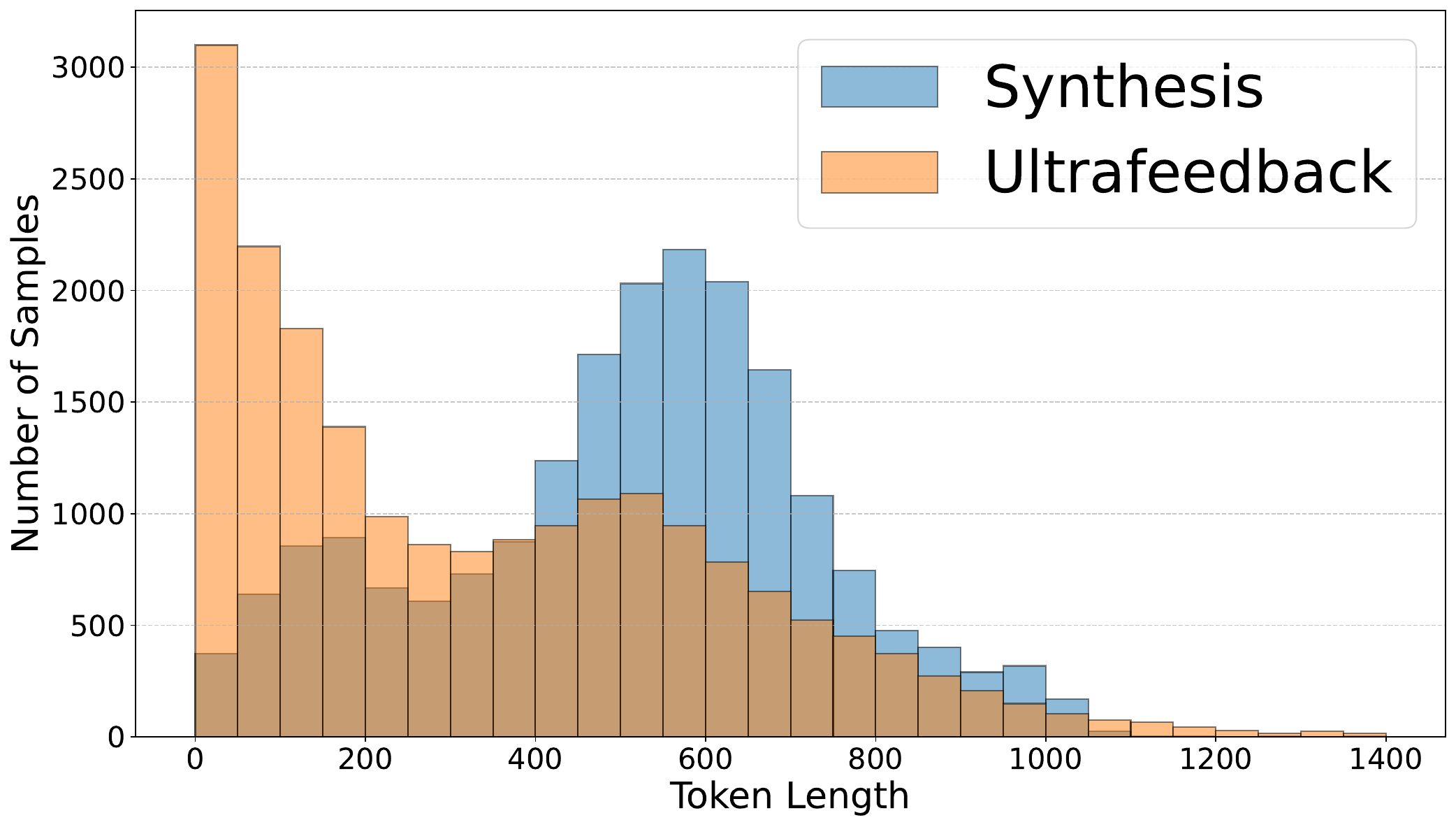}
    \end{tabular}
    }
    \caption{Distribution of prompt and response lengths.}
    \label{fig:length_distributions}
\end{figure}

The superior performance of SimPO may be attributed to the characteristics of our generated dataset, as shown in Figures \ref{fig:length_distributions}. Compared to externally labeled datasets, our synthetic dataset tends to generate shorter prompts (less than 200 words) and longer responses (mainly in the 400-800 words range), compared to the existing Ultrafeedback dataset. This characteristic makes SimPO's length normalization formula particularly effective in this context. Examples of the synthetic dataset are provided in Table \ref{dataset_analysis_example} in the Appendix.

\begin{table}[ht]
    \centering
    \caption{Experiment with different prompt sources.}
    \label{fig:repetition_rate_comparison}
    \scalebox{0.85}{
        \begin{tabular}{lcc}
            \toprule
            \textbf{Source} & \textbf{Win Rate (\%)} & \textbf{Repetition (\%)} \\
            \midrule
            UltraFeedback          & 55.84  & 0.0 \\
            \midrule
            Random  (Ours)      & 62.50  & 45.7 \\
            Persona (Ours)         & \textbf{72.30} & 0.7 \\
            \bottomrule
        \end{tabular}
    }
\end{table}

\paragraph{Different prompt sources.}

A key component of our method is persona-based prompt generation, which enhances the diversity of the generated prompts. Table \ref{fig:repetition_rate_comparison} illustrates the effect of using different prompt sources. When the synthetic dataset was generated with persona role-play, the model achieved a win rate (WR) of 72.30\% after training with it, along with a significantly lower prompt repetition rate of 0.7\% (only occurred when the model rejected generating prompts for certain personas). In contrast, when no persona role-play was used, the WR dropped to 62.50\%, and the repetition rate increased to 45.7\%.

These results emphasize the crucial role of persona-based prompt generation in improving model performance and reducing repetition. Notably, even in the presence of repetitive prompts, the model's performance remains strong. We attribute this robustness to the SAO, which allows the model to continue self-improving even when facing redundant prompts. The model remains capable of generating diverse responses, thanks to the next-token prediction mechanism and the temperature settings applied during generation.
For comparison, when the same SAO process was applied using randomly sampled prompts from UltraFeedback, the WR was only achieved 55.84\%. We believe this drop is due to the inclusion of prompts with mathematical and reasoning-based ground-truth data in the UltraFeedback dataset, which the model struggles to handle correctly, resulting in suboptimal optimization.

\begin{table}[t]
\centering
\caption{Comprehensive evaluation of feedback sources.}
\label{tab:sao_results}
\resizebox{0.92\columnwidth}{!}{
\begin{tabular}{lcc}
\toprule
\textbf{Feedback Source} & \textbf{LC (\%)} & \textbf{WR (\%)} \\
\midrule
\multicolumn{3}{l}{\textbf{Gemma-2-9B-it-SAO}} \\
\quad Random feedback & -- & 8.82 \\
\quad Length-Based feedback & -- & 3.29 \\
\quad ArmoRM feedback & -- & 41.43 \\
\quad GPT-4o feedback & 63.37 & 52.80 \\
\quad GPT-4o-mini feedback & \textbf{71.54} & 68.51 \\
\quad \textbf{Self-feedback(Ours)} & 70.28 & \textbf{74.04} \\
\quad \quad w/o Criterion 1 & 70.15 & 73.66 \\
\quad \quad w/o Criterion 2 & 70.22 & 73.98 \\
\quad \quad w/o Criterion 3 & 69.87 & 73.14 \\
\quad \quad w/o Criterion 4 & 69.43 & 72.67 \\
\bottomrule
\end{tabular}
}
\end{table}

\paragraph{Evaluation of feedback sources.}\label{sec:criteria_design}

To comprehensively evaluate the effectiveness of different feedback sources, we compared multiple feedback mechanisms. We tested several baseline approaches: Random (randomly selecting responses), Length-Based (preferring shorter responses), ArmoRM (using ArmoRM-Llama3-8B-v0.1 \citep{ArmoRM}), as well as feedback from external models (GPT-4o, GPT-4o-mini) and self-feedback. Additionally, we conducted an ablation study on the judging criteria used in our self-feedback template.

As shown in Table~\ref{tab:sao_results}, the results reveal clear patterns. For Gemma-2-9B-it, self-feedback significantly outperforms all alternatives, achieving WR of 74.04\% compared to GPT-4o feedback (52.80\%), GPT-4o-mini feedback (68.51\%), and ArmoRM feedback (41.43\%). Heuristic methods (Random: 8.82\%; Length-Based: 3.29\%) performed poorly, confirming the necessity of meaningful evaluation. The ablation study shows all criteria contribute to performance, with Criterion 4 being most critical (72.67\% when removed).


We further conducted a fine-grained quality analysis on responses selected by Gemma-2-9B-SAO feedback for given instructions, using five explicit metrics (0-10 scale), as shown in Table~\ref{tab:feedback_quality}. Self-feedback consistently achieves the highest average score (9.58), particularly excelling in Clarity (9.85) and Relevance (9.78).

\begin{table}[t]
\centering
\caption{Quality assessment of feedback from different sources in Gemma-2-9B-SAO dataset.}
\label{tab:feedback_quality}
\small
\begin{tabular}{lccc}
\toprule
\textbf{Metric} & \textbf{Self} & \textbf{GPT-4o-mini} & \textbf{GPT-4o} \\
\midrule
Correctness & 9.41 & 9.37 & 9.36 \\
Completeness & 9.45 & 9.42 & 9.34 \\
Clarity & 9.85 & 9.82 & 9.81 \\
Usefulness & 9.43 & 9.39 & 9.33 \\
Relevance & 9.78 & 9.76 & 9.76 \\
\midrule
Avg. Score & \textbf{9.58} & 9.55 & 9.52 \\
\bottomrule
\end{tabular}
\end{table}

\paragraph{Quality of generation vs. judging.}
\label{Judgers}
To disentangle the relative contributions of generation quality versus judging ability, we trained Gemma-2-9B-it on SAO datasets constructed using different combinations of generators and judges. Specifically, we used either Gemma-2-9B-it or LLaMA-3-8B-instruct as the generator (producing response pairs for prompts generated by Gemma-2-9B-it) and as the judge (selecting preferred responses from these pairs).

As shown in Table~\ref{tab:generation_vs_feedback}, judging ability has a substantially greater impact on alignment performance than generation quality. When Gemma serves as the judge, both configurations achieve strong results: Gemma-generated + Gemma-judged yields the best performance (LC: 70.28\%, WR: 74.06\%), while LLaMA-generated + Gemma-judged also performs well (LC: 67.30\%, WR: 72.73\%). In contrast, when LLaMA serves as the judge, performance drops significantly regardless of the generator—even Gemma-generated responses with LLaMA judging achieve only LC: 58.39\% and WR: 34.35\%, barely improving over the vanilla baseline.

These results demonstrate that judging quality is the critical factor in alignment effectiveness. A model with strong judging ability can effectively utilize preference data regardless of its generation source, while weak judging ability limits performance even with high-quality generated data. This finding validates our core claim that leveraging a model's intrinsic judging capability can effectively drive self-improvement in alignment.

\begin{table}[t]
\centering
\caption{Performance of Gemma-2-9B-it under different generation and judging configurations. Generator produces response pairs; judge selects preferred responses.}
\label{tab:generation_vs_feedback}
\resizebox{0.8\columnwidth}{!}{
\begin{tabular}{lcc}
\toprule
\textbf{Generator + Judge} & \textbf{LC (\%)} & \textbf{WR (\%)} \\
\midrule
Vanilla Gemma-2-9B-it & 56.50 & 39.30 \\
\midrule
Gemma + LLaMA & 58.39 & 34.35 \\
LLaMA + LLaMA & 59.32 & 34.41 \\
LLaMA + Gemma & 67.30 & 72.73 \\
Gemma + Gemma & \textbf{70.28} & \textbf{74.06} \\
\bottomrule
\end{tabular}
}
\end{table}

\paragraph{Multi-judge evaluation.}

To further validate the robustness of SAO improvements and address potential evaluation bias, we evaluated our models using multiple independent judge models beyond the default GPT-4 and Qwen2-72B-instruct judges in AlpacaEval 2.0. Specifically, we employed three diverse state-of-the-art LLM evaluators: LLaMA-3.3-70B, LLaMA-3.1-70B, and Qwen-2.5-72B to evaluate models trained with self-feedback and LLaMA-feedback.

As shown in Table~\ref{tab:multi_judge}, Gemma-2-9B-it with self-feedback demonstrates consistent and substantial improvements across all three judges. With LLaMA-3.3-70B as the evaluator, WR increases from 45.22\% to 73.11\%; with LLaMA-3.1-70B, from 42.66\% to 69.50\%; and with Qwen-2.5-72B, from 42.61\% to 76.09\%. The LC metric also shows steady gains across all evaluators.

In contrast, SAO trained with LLaMA-feedback consistently underperforms the self-feedback approach, with WR ranging from 32.67\% to 38.39\% across different judges. These results confirm that the performance gains from our SAO method are robust and generalize across different evaluation perspectives, demonstrating genuine alignment improvements rather than judge-specific overfitting.

\begin{table}[t]
\centering
\caption{Multi-judge evaluation on AlpacaEval 2.0 showing consistent improvements across different evaluators.}
\label{tab:multi_judge}
\resizebox{1.0\columnwidth}{!}{
\begin{tabular}{llcc}
\toprule
\textbf{Judge} & \textbf{Model Variant} & \textbf{LC (\%)} & \textbf{WR (\%)} \\
\midrule
\multirow{3}{*}{LLaMA-3.3-70B} 
& Gemma-2-9B-it & 64.19 & 45.22 \\
& \quad + SAO (self) & \textbf{68.26} &  \textbf{73.11} \\
& \quad + SAO (LLaMA) & 59.96 & 36.77 \\
\midrule
\multirow{3}{*}{LLaMA-3.1-70B} 
& Gemma-2-9B-it & 58.46 & 42.66 \\
& \quad + SAO (self) &  \textbf{64.54} &  \textbf{69.50} \\
& \quad + SAO (LLaMA) & 58.65 & 38.39 \\
\midrule
\multirow{3}{*}{Qwen-2.5-72B} 
& Gemma-2-9B-it & 61.33 & 42.61 \\
& \quad + SAO (self) &  \textbf{71.89} &  \textbf{76.09} \\
& \quad + SAO (LLaMA) & 55.82 & 32.67 \\
\bottomrule
\end{tabular}
}
\end{table}
\section{Conclusion}

In this paper, we propose a self-alignment optimization framework for aligning LLMs with human preferences using only self-generated synthetic data. Our approach enables model itself to generate and judge instruction-preference pairs, eliminating the need for costly external annotations. Through comprehensive experiments, we demonstrate that our framework achieves substantial performance gains on subjective evaluation benchmarks while maintaining or improving performance on downstream objective tasks. We further demonstrate promising scalability through dataset size scaling and iterative optimization, where improved models generate higher-quality preference data for subsequent rounds. We hope SAO provides a practical path toward self-improvement and unlocking the latent capabilities of LLMs.

\section{Limitations}
While our experimental results are promising, this study is constrained by the use of models smaller than 10 billion parameters due to resource limitations. We anticipate that scaling the SAO framework to larger models could yield even greater performance enhancements. Additionally, although our approach has demonstrated effectiveness with simple prompt templates, investigating more complex templates may provide further improvements. Future research should address these areas to fully leverage the potential of the SAO framework.


\bibliography{arr_conference,sao}

\newpage
\appendix
\onecolumn

\section{Impact Statements}
The introduction of the SAO framework offers a valuable solution to the challenges of fine-tuning LLMs without extensive external supervision. This approach can significantly reduce the manual effort and time required for model training, thereby enhancing accessibility to NLP technologies for individuals and organizations with limited resources. However, it is crucial to exercise caution with this self-improvement framework, as it relies entirely on self-synthesized datasets, which may lead to the generation of inaccuracies or hallucinations in certain cases.

\section{Experiments on Small-Sized LLMs}
Interestingly, when we extend our SAO framework to smaller models (1B-3B), we observe that the initial subjective and alignment capabilities of the backbone model significantly impact the results. For instance, Llama-3.2-Instruct-3B shows a performance degradation after applying SAO, while Gemma-2-2B-it demonstrates notable improvements. We hypothesize that weaker models may lack sufficient judgment ability to effectively conduct self-alignment.

\begin{table}[ht]
    \centering
    \caption{Performance of small language models on AlpacaEval 2.0 using Qwen2-72B-Instruct as the judge.}
    \label{tab:result-alpaca-qwen-small}
    \resizebox{0.5\linewidth}{!}{  
        \begin{tabular}{l c c c}
            \toprule
            \multirow{2}{*}{Model} & \multicolumn{3}{c}{AlpacaEval 2.0} \\
            \cmidrule(lr){2-4}
            & LC (\%) & WR (\%) & STD \\
            \midrule
            Llama-3.2-Instruct-3B          & 21.87 & 24.97 & 1.52 \\
            Llama-3.2-Instruct-3B-SAO      & 20.08\,(-1.79) & 22.05\,(-2.92) & 1.46 \\
            \hline
            Gemma-2-2B-it                  & 38.08 & 41.74 & 1.74 \\
            Gemma-2-2B-it-SAO              & 46.18\,(+8.10) & 49.13\,(+7.39) & 1.76 \\
            \bottomrule
        \end{tabular}
    }
\end{table}

\section{Repetition in Generated Prompts Under Random Generation}

\begin{table*}[ht]
\centering
\small  
\renewcommand{\arraystretch}{1.0}  
\caption{Top 5 repeated randomly generated prompts without persona role-play in 10k synthesized dataset.}
\begin{tabularx}{0.9\textwidth}{>{\raggedright\arraybackslash}p{0.7\textwidth} r}
\hline
\textbf{Prompt} & \textbf{Count} \\
\hline
Summarize the main plot points of the novel "Pride and Prejudice" by Jane Austen, focusing on the development of the relationship between Elizabeth Bennet and Mr. Darcy. & 267 \\
Summarize the main plot points of the novel "Pride and Prejudice" by Jane Austen, focusing on the relationship between Elizabeth Bennet and Mr. Darcy. & 258 \\
Write a short story (around 500 words) about a young inventor who creates a device that can translate animal languages, but struggles with the ethical implications of their invention. & 178 \\
Summarize the main plot points of the novel *Pride and Prejudice* by Jane Austen, focusing on the relationship between Elizabeth Bennet and Mr. Darcy. & 114 \\
Summarize the main plot points of the novel *Pride and Prejudice* by Jane Austen, focusing on the development of the relationship between Elizabeth Bennet and Mr. Darcy. & 105 \\
\hline
\end{tabularx}
\label{repeated_prompts_without_persona}
\end{table*}

\newpage
\section{Prompt Template}

\begin{figure*}[ht]
\centering
\begin{tcolorbox}[
  colback=green!5!white,
  colframe=green!50!black,
  width=0.95\textwidth,
  arc=2mm,
  title=Persona Instruction Example
]
Guess a prompt that the following persona may ask you to do: \\

A Political Analyst specialized in El Salvador's political landscape.

\vspace{0.5em}

\textbf{Note:} \\
1. The prompt should be informative and specific. \\
2. Your output should start with "User prompt:"
\end{tcolorbox}

\vspace{0.1em}

\begin{tcolorbox}[
  colback=blue!5!white,
  colframe=blue!50!black,
  width=0.95\textwidth,
  arc=2mm,
  title=Pair-wise Response Ranking
]
You are an impartial judge. Your task is to rank two answers to a given prompt based on their quality.

\vspace{0.5em}

\textbf{Prompt:} \{prompt\} \\

\textbf{Response 1:} \texttt{<Response 1>} \{response 1\} \texttt{</Response 1>} \\

\textbf{Response 2:} \texttt{<Response 2>} \{response 2\} \texttt{</Response 2>} \\

\vspace{0.5em}

Please carefully read each response and evaluate them based on the following criteria: \\
1. Relevance and specificity to the prompt \\
2. Accuracy and correctness of information \\
3. Completeness and comprehensiveness \\
4. Clarity and understandability

\vspace{0.5em}

Then, rank these two responses from best to worst. 
You must output your ranking strictly in the following format: 
\textbf{ranking: X \textgreater{} Y}, where X and Y represent one of 1 or 2, without repetition.

\vspace{0.5em}

Remember, you must output a complete ranking including both options. Now, please provide your ranking:
\end{tcolorbox}

\caption{The top box displays the persona instruction prompt, which directs the LLM to generate a specific prompt based on a given persona. The bottom box illustrates the pair-wise response ranking prompt, instructing the LLM to compare and rank responses based on specific criteria modified from \citet{shen2024boostingrewardmodelpreferenceconditional}.}
\label{fig:eval_prompt}
\end{figure*}

\newpage
\section{Synthesized Examples}

\begin{table*}[ht]
    \centering
    \renewcommand{\arraystretch}{1.5} 
    \setlength{\tabcolsep}{4pt} 
    \resizebox{0.9\linewidth}{!}{%
    \begin{tabular}{|p{0.2\textwidth}|p{0.75\textwidth}|} 
    \hline
    \textbf{Input Persona} & A newly hired general counsel at TurpCo Industries \\
    \hline
    \textbf{Generated Prompt} & As a newly hired General Counsel at TurpCo Industries, I need to understand the legal landscape of our operations. Could you provide a detailed report on the major legal risks and compliance issues our company might face in the next fiscal year, including any potential changes in legislation that could affect our business? Additionally, please include recommendations for mitigating these risks and ensuring our compliance with all relevant laws and regulations. \\
    \hline
    \textbf{Response 1} & \textbf{Executive Summary:} \par
        As the General Counsel of TurpCo Industries, it is essential to understand the legal risks and compliance issues that our company may face in the next fiscal year. This report provides an overview of the major legal risks and compliance issues that our company may face, including potential changes in legislation that could affect our business. The report also includes recommendations for mitigating these risks and ensuring our compliance with all relevant laws and regulations. \par
        \textbf{Legal Risks and Compliance Issues:} \par
        1. \textbf{Data Privacy and Cybersecurity:} As the world becomes increasingly digital, data privacy and cybersecurity risks have become more prevalent. Our company must ensure that we comply with all relevant data protection laws and regulations, such as the General Data Protection Regulation (GDPR) in Europe and the California Consumer Privacy Act (CCPA) in the United States. We must also have robust cybersecurity measures in place to protect our company's and our customers' data...... \par \\

    \hline
    \textbf{Response 2} & \textbf{Subject: Legal Risk Assessment and Compliance Report for TurpCo Industries} \par
        Dear [Recipient's Name], \par
        I am pleased to present a comprehensive report on the major legal risks and compliance issues that TurpCo Industries may face in the upcoming fiscal year. This report will also address potential legislative changes that could impact our operations, along with recommendations for risk mitigation and ensuring legal compliance. \par
        1. \textbf{Overview of Legal Risks and Compliance Issues:} \par
           a. \textbf{Environmental Regulations:} \par
              - \textbf{Risk:} TurpCo Industries' manufacturing processes may be subject to stringent environmental regulations, including those related to emissions, waste disposal, and water usage. \par
              - \textbf{Potential Changes:} The Environmental Protection Agency (EPA) is considering updating regulations to reduce greenhouse gas emissions and tighten waste disposal standards..... \par \\

    \hline
    \textbf{Ranking} & 2 $>$ 1 \\
    \hline
    \end{tabular}
    }
    \caption{Example of synthetic dataset.}
    \label{dataset_analysis_example}
\end{table*}

\newpage
\subsection{Model Outputs Before and After SAO Training}

\begin{table*}[ht]
    \centering
    \scalebox{0.83}
    {
    \begin{tabular}{|p{0.9\textwidth}|}
        \hline
        \textbf{Instruction} \\ 
        \hline
        What are the names of some famous actors that started their careers on Broadway? \\ 
        \hline
        \textbf{Gemma2-9B-it Output} \\ 
        \hline
        Here are some famous actors who started their careers on Broadway: \\
        
        **Legends:** \\
        * **Meryl Streep:** Made her Broadway debut in 1975 and won a Tony Award for "The Cherry Orchard." \\
        * **Denzel Washington:** Won a Tony Award for his performance in "Fences" and has starred in many other Broadway productions. \\
        * **Al Pacino:** Rose to fame on stage before transitioning to film, winning a Tony Award for "Does a Tiger Wear a Necktie?" \\
        * **Robert De Niro:** Made his Broadway debut in 1964 and has since returned for several productions. \\
        * **James Earl Jones:** Known for his iconic voice, he won a Tony Award for "Fences" and has been a mainstay on Broadway for decades. \\

        **More Recent Stars:** \\
        * **Lin-Manuel Miranda:** Created and starred in the smash hit "Hamilton," winning multiple Tony Awards for his work. \\
        * **Andrew Garfield:** Earned Tony Award nominations for "Death of a Salesman" and "Angels in America." \\
        * **Scarlett Johansson:** Made her Broadway debut in "A View from the Bridge" and has since returned for other productions. \\
        * **Nathan Lane:** A Tony Award-winning actor known for his comedic roles in musicals like "The Producers" and "Guys and Dolls." \\
        * **Hugh Jackman:** Known for his role as Wolverine, he has also starred in several Broadway musicals, winning a Tony Award for "The Boy From Oz." \\

        This is just a small sample of the many talented actors who have honed their craft on Broadway. Let me know if you'd like to explore actors from a specific genre or era! \\ 
        \hline

        \textbf{Gemma2-9B-it-SAO Output} \\ 
        \hline
        Here's a list of famous actors, spanning generations and genres, who honed their craft and gained initial recognition on Broadway: \\

        **Stage Legends \& Classics:** \\
        * **Marlon Brando, James Dean, Montgomery Clift:** Pioneered Method acting, rose to fame through iconic stage productions before Hollywood. \\
        * **Humphrey Bogart, Katharine Hepburn, Laurence Olivier, Ingrid Bergman:** Established themselves as stage stars before transitioning to film. \\
        * **Judy Garland, Ethel Merman, Carol Channing, Yul Brynner:** Musical theatre icons who built enduring careers on Broadway. \\
        * **Angela Lansbury, Julie Andrews, Audra McDonald:** Multiple Tony Award winners with long, distinguished Broadway careers. \\

        **Contemporary Stars:** \\
        * **Denzel Washington, Viola Davis, Nathan Lane, Patrick Stewart:** Highly acclaimed actors with both stage and screen success. \\
        * **Tom Hanks, Meryl Streep, Daniel Day-Lewis, Michelle Williams:** Oscar winners who began their careers in theatre. \\
        * **Idina Menzel, Lin-Manuel Miranda, Andrew Garfield, Jake Gyllenhaal:** Known for both stage and screen, often returning to Broadway. \\
        * **Hugh Jackman, Bradley Cooper, Emma Stone, Scarlett Johansson:** Made their mark on Broadway before achieving major film success. \\

        **Notable Mentions:** \\
        * **Robert De Niro, Al Pacino, Christopher Walken:** Strong stage backgrounds, though perhaps less known for their Broadway work specifically. \\

        This list is by no means exhaustive, but it highlights the rich history of actors who found their footing and developed their skills on the Broadway stage. \\ 
        \hline
    \end{tabular}
    }
    \caption{Outputs from Gemma2-9B-it and Gemma2-9B-it-SAO models.}
    \label{tab:actors_broadway}
\end{table*}

\end{document}